\documentclass{article}
\usepackage{graphicx}
\graphicspath{ {./figures/} }

    \usepackage[preprint]{neurips_2022}

\usepackage[utf8]{inputenc} 
\usepackage[T1]{fontenc}    
\usepackage{hyperref}       
\usepackage{url}            
\usepackage{booktabs}       
\usepackage{amsfonts}       
\usepackage{nicefrac}       
\usepackage{microtype}      
\usepackage{xcolor}         
\usepackage{caption}
\usepackage{subcaption}

\title{Open Arms: Open-Source Arms, Hands \& Control}

\author{%
  David Hanson \\
  Hanson Robotics Ltd\\
  \texttt{david@hansonrobotics.com} \\
  \And
   Alishba Imran \\
   Hanson Robotics Ltd \\
   \texttt{alishbai734@gmail.com} \\
   \And
   Gerardo Morales \\
   Hanson Robotics Ltd \\
   \texttt{gerardo@hansonrobotics.com} \\
   \And
   Vytas Krisciunas \\
   Hanson Robotics Ltd \\
   \texttt{vytas@hansonrobotics.com} \\
   \And
   Aditya Sagi \\
   Indian Institute of Science \\
   \texttt{adityavarma@iisc.ac.in } \\
   \And
   Aman Malali \\
   ARTPARK \\
   \texttt{aman@artpark.in} \\
   \And
   Rushali Mohbe \\
   ARTPARK \\
   \texttt{rushali@artpark.in} \\
   \And
   Raviteja Upadrashta \\
   ARTPARK \\
   \texttt{raviteja.upadrashta@gmail.com} \\
}

\begin{document}

\maketitle
\begin{abstract}
  Open Arms is a novel open-source platform of realistic human-like robotic hands and arms hardware with 28 Degree-of-Freedom (DoF), designed to extend the capabilities and accessibility of humanoid robotic grasping and manipulation. The Open Arms framework includes an open SDK and development environment, simulation tools, and application development tools to build and operate Open Arms. This paper describes these hands' controls, sensing, mechanisms, aesthetic design, and manufacturing and their real-world applications with a teleoperated nursing robot. From 2015 to 2022, the authors have designed and established the manufacturing of Open Arms as a low-cost, high functionality robotic arms hardware and software framework to serve both humanoid robot applications and the urgent demand for low-cost prosthetics, as part of the Hanson Robotics Sophia Robot platform. Using the techniques of consumer product manufacturing, we set out to define modular, low-cost techniques for approximating the dexterity and sensitivity of human hands. To demonstrate the dexterity and control of our hands, we present a Generative Grasping Residual CNN (GGR-CNN) model that can generate robust antipodal grasps from input images of various objects in real-time speeds ($\sim$22ms). We achieved state-of-the-art accuracy of 92.4\% using our model architecture on a standard Cornell Grasping Dataset, which contains a diverse set of household objects.
\end{abstract}

\section{Introduction}

Today, multi-DOF robot hands can already accomplish various tasks that only humans can perform otherwise. However, the development of low-cost human-like robot hands and the precise manipulation control of everyday objects with such hands remains a challenge. 

Most existing robotic hands are costly, difficult to maintain because of tendon-driven actuators, or do not provide precise control. The TWENDY-ONE Hand [1] has stable manipulation skills, but its usage is limited in industrial applications due to the high development and maintenance cost of highly sophisticated actuators and sensors. Similarly, the multi-fingered Shadow Dexterous Hand can accomplish complex tasks [2], but costs about \$100,000, and the tendon-driven actuators make the maintenance difficult. A more cost-effective hand is the Wonik’s Allegro Hand [3], which is relatively low-cost (\$15,000) but does not provide precise position or force control. Open Bionics developed the Hero arm [4] as a lower-cost 3D printed alternative to contemporary prosthetics but these prosthetics lack intelligent controls. We note opportunities to improve on these existing projects by adding additional naturalism to the design, with improved actuation, sensing, intelligent controls, and scalable consumer-grade manufacturing to reduce costs while achieving high quality. 

In this paper, we describe how our novel robotic arms can be designed such that the fabrication of the robotic hand is fast, the cost of the modification and maintenance is cheap, and the control of the robotic hand is accurate. We present the design and construction of novel robotic arms as a new platform, integrating new hardware design and the controls of the hands and arms using a cognitive AI framework involving GGR-CNN for accurate grasping. 

The Open Arms include novel robot arms built with a humanlike mechanical configuration and aesthetic, with 28 DoF, touch sensors, and series elastic actuators. The arms and hands have many improvements such as retraction and elevation of the shoulders, a reduction in its weight, as well as more grip strength that can even lift a 600g bottle, pressure sensors in the fingertips and position sensors in the finger joints, giving you the option to receive feedback when manipulating objects [5], [6]. 

The main contributions of this paper can be summarized as follows:
\begin{itemize}
    \item The Open Arms is a robotics framework that pushes forward realistic design with improved actuation, sensing, controls, and scalable manufacturing methods to reduce the costs of 28 DoF human-like robotic hands and arms. 
    \item The platform includes an open SDK and development environment to build, customize, and operate Open Arms. 
    \item We propose a Generative Grasping Residual CNN (GGR-CNN) model for the hands that predicts, plans, and performs antipodal grasps for the objects in the camera's field of view.
    \item We evaluate our grasping model on the publicly available Cornell grasping dataset and achieve state-of-the-art accuracy of 92.4\%.
    \item We demonstrate the real-world use of our arms and hands by testing and outlining the design of a teleoperated nursing robot.  
\end{itemize}

The resulting framework is an integral part of the Sophia Utility Platform (SUP), which is being used with ongoing research in the authors’ work with team AHAM, an ANA Avatar Xprize effort towards human-AI hybrid telepresence. The uses of the work extend across domains and include arts, social human-robot interaction, and targeting more general co-bot applications. These results are available to test on the broadly released Sophia Utility Platform (SUP), including Open Arms, for users to try and extend.

\section{Background}
\subsection{Open Arms, Part of Sophia Utility Platform: An Integrative Platform for Embodied Cognition }

The Open Arms were developed as part of our latest designed framework for human-like embodied cognition, called Sophia Utility Platform (SUP). 

While some cognitive robotics and AI frameworks exist, such as iCub [7], SOAR [8], ACT-R [9], and Open Cog [10], none provide the full, integrated capabilities needed for the pursuit of humanlike embodied cognition, including low-cost mass-manufactured standard hardware for hands and arms, expressive face, open interfaces, and a diversity of features and extensions. The SUP provides many features, including an expressive face, Open Arms, a mobile base, and a feature-rich AI SDK on a new mass-manufacturable hardware and software framework for cognitive robotics. With SUP, we strive to architect a humanlike agent with embodiment and cognitive AI tools for an agent to learn its environment using cutting-edge multimodal learning methods in vision, speech, grasping and manipulation, locomotion, and social interactions with human users.  We combine various neural frameworks with a rules-based controller governing an ensemble verbal and non-verbal dialogue model that is original to us. Using basic emotional parameters to drive the agent’s goal pursuit, making the agent more useful physical actions like the Open Arms and hands control while also achieving socially intelligible and appealing human-robot interactions. 

\section{Open Arms Robotic Hardware Design}
Open Arms have 28 DoF, and the hardware is designed to simulate the action, appearance, and capabilities of the human hands and arms. Each arm includes 2 DoF for the shoulder joint (pitch and roll), 1 DoF for the shoulder or elbow yaw between the shoulder and the arm, 1 DoF for the elbow (pitch), 1 DoF for the forearm (yaw), and 2 DoF for the wrist (pitch, roll). Each hand includes 2 DoF for the thumb, 1 DoF for the finger spread, and 4 DoF for each of the other fingers controlled by a single actuator.

The hands and arms have sensing, including force, position, temperature feedback, compliant series elastic actuators in the fingers, and touch sensing in the fingertips. The arms were designed to be modular so that the hands and forearms could also be used in low-cost prosthetics and could be easily changed and upgraded. The first iteration of the designs was completed in 2014, intended both for research and for mass-production as part of the Sophia Utility Platform line of social robots for general use, including human-AI interactions and collaborative robotics (co-botics). Early versions were released under Creative Commons open license in 2016 and 2017. From 2020 to 2022, we completed improved Beta designs and initiated the manufacturing of the Open Arms in volume units. These latest versions are now released under the General Public License and the designs are available for any developer to use freely. The design files for the arms are attached as supplementary material.

\begin{figure}
  \centering
  \includegraphics[width=\linewidth]{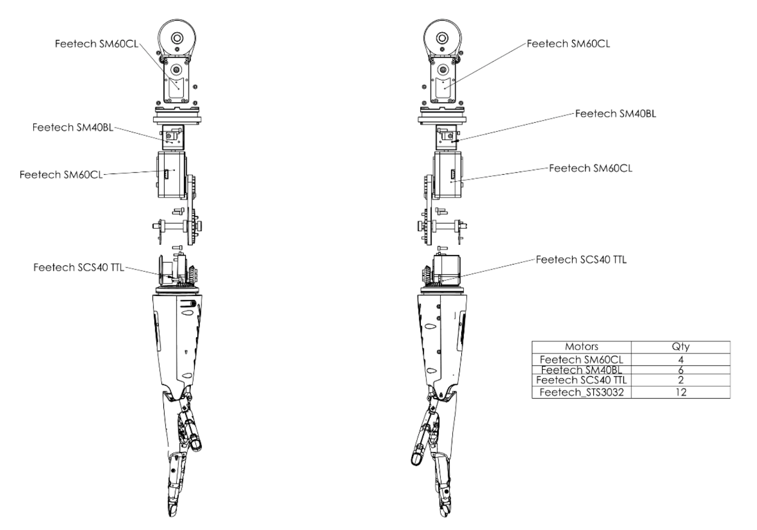}
  \caption{Servos and design of the Open Arms}
\end{figure}

\subsection{Mechanical Design}
Mechanically, we developed our robotic arms to suit the character and collaborative robotic needs. We used the dimensions from a medium-sized woman's arms as constraints and quality components integrated within the robots. We try to have feedback in all the DOF axes where possible. The principal structure of the arms is made of CNC Aluminum 6061, and complicated manufactured parts are made of Nylon SLS PA 2200, black colour for aesthetic. Feetech servos are used for controlling and for feedback.  Due to the size limitation, each hand uses our versions of Feetech servos with Pololu Maestro Controller 24ch. For feedback, there are pressure sensor pads and potentiometers in each finger and wrist. The goal of our arms is for gesticulation when she speaks and for object grasping and manipulation. 

The shoulders on the torso have new features: elevation-depression (22° range) and retraction-protraction (23° range) using 2 Actuonix+L12-30 linear actuators on each side motors are connected with universal joints. These characteristics in the shoulders aggregate us with more realistic movements and better gestures.

The robotic arms have 7 DOF including:
\begin{itemize}
    \item Shoulder pitch (180° range).
    \item Shoulder roll (180° range), which are driven by a pulley and timing belt with a 2:1 ratio using Feetech SM60CL motors each.
    \item Shoulder yaw (120° range) is driven by gear with a 2.917:1 ratio with Feetech SM40BL motor.
    \item Elbow uses the same belt pulley with a 2:1 ratio (range 127°) and Feetech SM60CL motor.
    \item Wrist yaw (180° range) uses bevel gears with a 1.8:1 ratio with Feetech SCS40 TTL motor.
    \item Finally, wrist roll (40° range) and pitch (45° range) are powered by two Feetech SM40BL motors with universal joints simultaneously.
\end{itemize}
The hand has 6 DOF, all with small Feetech STS3032 motors to drive the 5 fingers and the thumb roll with metal wires. The hand closes when the motors pull the wires and opens when they release it along with the help of torsion springs that are in the joints of each finger. 

The way to connect the arm and send the information to the computer is through the Feetech FE-URT1 converter, which receives communication from the SM40BL and SM60CL motors connected in series via RS485 protocol and the rest of the STS3032 and SCS40TTL motors connected in series via TTL protocol.

\begin{figure}
  \centering
  \begin{subfigure}[b]{0.4\linewidth}
    \includegraphics[width=\linewidth]{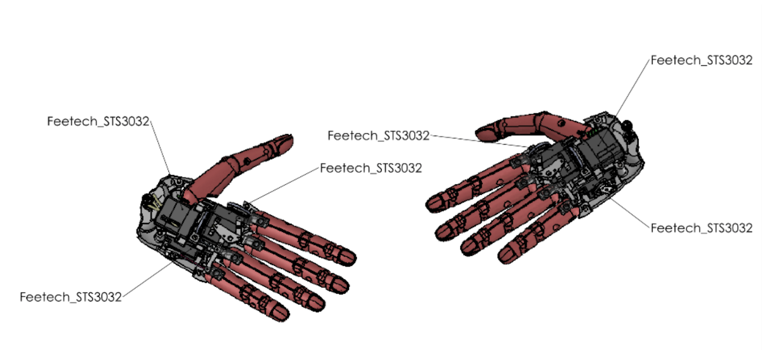}
  \end{subfigure}
  \begin{subfigure}[b]{0.4\linewidth}
    \includegraphics[width=\linewidth]{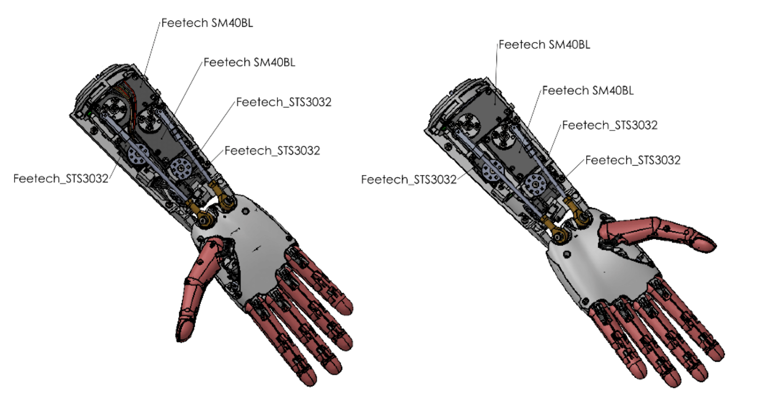}
  \end{subfigure}
  \caption{Servos on our open hands}
  \label{fig:coffee}
\end{figure}

\subsection{Arm Axes}
The arms use Feetech servos, and the selection of those servos is determined by the weight, torque and space limitations. The torque of each Axes was calculated in Solidworks with the center of mass and the weight.

The features of Feetech servos are: Smart actuator with fully integrated DC Motor + Reduction Gearhead + Encoder + Controller + Driver. Functions include precise control, PID control, 360 degrees of position control, and high-speed communication. 

\subsection{Hand Axes}
Due to space limitations, there are 3 Feetech servos in hand and 3 more in the forearm, which control the fingers. For the feedback, each fingertip has an internal pressure sensor pad measuring 500gr; the sensor is activated through a Nylon-rubber fingertip cover.  

There are 16 10K ohms potentiometers for measuring positions: index, middle, ring, pinky, thumb fingers, thumb roll, wrist pitch and wrist roll. There are torsion springs in fingers joints in one direction for safety use.

\subsection{Electrical}
The power of the Arm is 13.8V, stepped down from 24V for SM40BL and SM60CL motors and power of 6V for STS3032 and SCS40TTL motors, and the communication is through the Feetech FE-URT1 converter, which receives communication from the SM40BL and SM60CL motors connected in series via RS485 protocol and the rest of the STS3032 and SCS40TTL motors connected in series via TTL protocol. The sensors like pressure sensors and potentiometers are connected through two 24 channel Pololu Maestro-microcontrollers in each forearm.

\section{Open Arms, Grasping: Generative Grasping Residual CNN (GGR-CNN)}

\subsection{Deep Learning Methods Using CNNs for Grasping}
Deep learning techniques have enabled big advancements in robotic grasping for unknown items. These approaches typically use variations of Convolutional Neural Network (CNN) architectures designed for object recognition [11, 12], and in most cases sample and rank grasp candidates individually using a CNN [13, 14]. Once the best grasp candidate is determined, a robot arm executes the grasp open-loop which requires very precise control of the robot and a completely static environment. This results in long computation times in the order of a second to tens of seconds. 

To overcome these challenges, we leverage the Generative Grasping Convolutional Neural Network (GG-CNN) [15] to directly generate an antipodal grasp pose and quality measure for every pixel in an input depth image. This model does not rely on sampling of grasp candidates, but rather directly generates grasp poses on a pixel-wise basis. However, the original GG-CNN method achieves an average grasping pose accuracy of only 83.83\%. To increase the accuracy of this model, we integrate residual networks as many state-of-the-art results have been obtained by using residual networks for deeper architectures [16, 17]. The best grasp pose is predicted as a grasp rectangle calculated by choosing the best grasp from multiple grasp probabilities. Multiple grasp rectangles can be predicted for multiple objects from the output of GGR-CNN in one shot which leads to faster computation. Through this approach and improved data augmentation methods, we are able to increase the accuracy from 73.0\% using the GG-CNN to 92.4\% on the Cornell grasping dataset using our GGR-CNN model.

\subsection{Grasp Representation }
Similar to other CNN based grasping methods, the GG-CNN defines its grasps perpendicular to a 2D axis with the following equation:

\label{equations}
\begin{equation}
\label{eqn1}
    g=(p,\phi,w,q) 
\end{equation}

Here \(p=(x,y,z)\) indicates the pose of the gripper as the Cartesian coordinates \((x, y)\) of the gripper’s centre relative to an image, $\phi$ indicates the rotation of the gripper around the z-axis, $w$ is the width of the gripper and $q$ is the grasp quality score. The final grasp is selected based on the q parameter since it represents the success probability of the grasp and indicates the best grasp.

We detect a grasp from an image \(I=R^{n\times h\times w}\) with height $h$ and width $w$. The final grasping pose has the output:

\begin{equation}
\label{eqn2}
    \tilde{g}=(s,\tilde{\varphi},\tilde{\omega},q)
\end{equation}

In equation (2), $s=u,v$ corresponds to the center of grasp pose $g$ in image coordinates grasping pose, $\tilde{\varphi}$ is the rotation angle in camera’s frame of reference which is represented as a value in the range $[-\frac{\pi}{2}, \frac{\pi}{2} ]$, $\tilde{\omega}$ is the required width in image coordinates and $q$ is the same scalar which represents the grasp quality score. $q$ is indicated as a value between 0 and 1 where a value closer to 1 indicates a greater chance of grasp success.

To execute on a grasp pose $\tilde{g}$ which is obtained in the image space on a robot, the grasping pose can be converted into the grasping pose g defined in the world coordinate system:

\begin{equation}
\label{eqn3}
g=t_{RC}(t_{CI}(\tilde{g}))
\end{equation}

Where, $t_{CI}$ is a transformation that converts image space into camera’s 3D space using the intrinsic parameters of the camera, and $t_{RC}$ converts from the image coordinate system into the robot space in the world using the camera pose calibration value. 

We can scale this to define multiple grasps in an image using the following:

\begin{equation}
\label{eqn4}
G=(\Phi,W,Q)\in R^{3\times H\times W}
\end{equation}

$\Phi$, $W$, $Q$ represent the three images of grasp angle, grasp width and grasp quality score calculated at every pixel of an image, respectively.

\section{Neural Network Architecture: Generative Grasping Residual CNN}

\begin{figure}
  \centering
  \includegraphics[width=\linewidth]{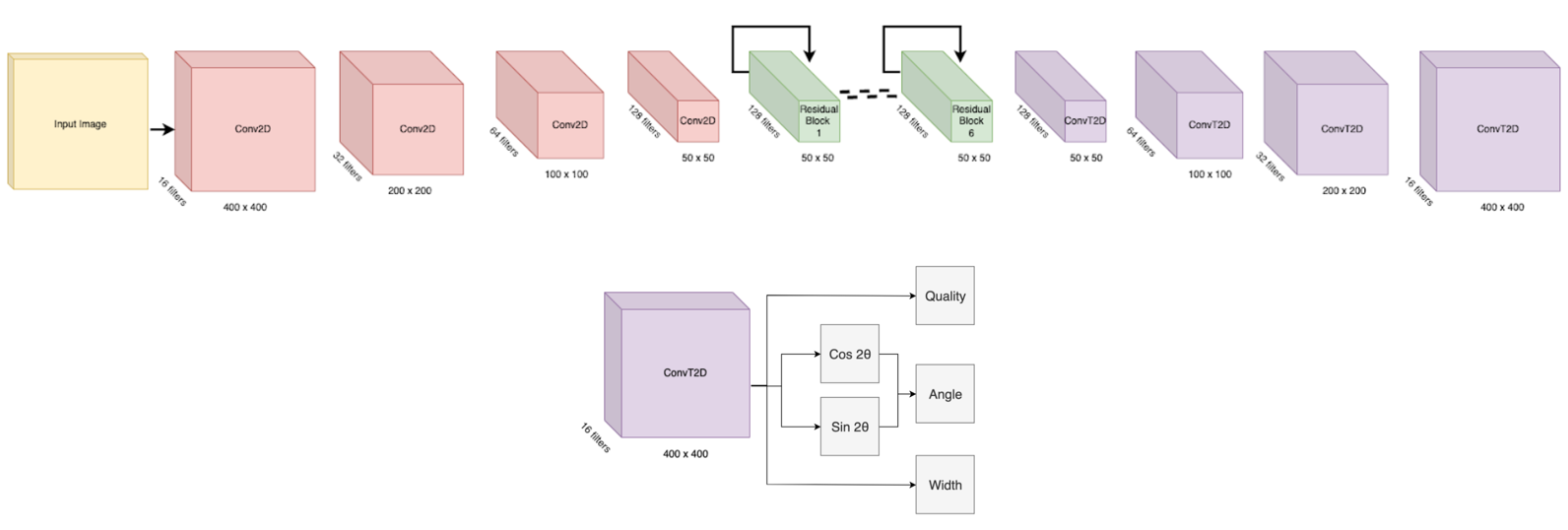}
  \caption{The GGR-CNN architecture includes the encoder, decoder and output of model which can be seen from the last transpose convolutional layer.}
\end{figure}

\subsection{Grasp Dectection Pipeline}
The depth image is first cropped to a square and scaled to suit the input of the GG-CNN. The depth image is inpainted to obtain a depth representation. The 400 × 400 processed input image is fed into the GGR-CNN. The model generates three images as grasp angle, grasp width, and grasp quality score as the output. From these three output images, the grasp rectangles are then inferred. 

\subsection{Architecture}
The full architecture can be seen in Figure 3. The input image is fed to the GGR-CNN, where it is passed through 4 convolutional layers, followed by 6 residual layers and then passed through convolution transpose layers to generate four images. These four output images include the grasp quality score, required angle (cos 2$\Phi$  and sin 2$\Phi$ ), and the required width of the end effector. 
The GGR-CNN’s first layer is 400 × 400 layer with 16 filters. The second layer is a 200 × 200 layer with 32 filters, followed by a third 100 × 100 layer with 64 filters and a fourth layer that is 50 × 50 layer with 128 filters. The output of the convolutional layer is then fed into 6 residual layers. Using residual layers enables us to better learn the identity functions by using skip connections. After passing the image through these convolutional and residual layers the size of the image is reduced to 50 × 50. To still retain spatial features of the image, we up-sample the image by using the convolution transpose operation. This allows us to obtain the same size of the image at the output as the input size. 

\section{Evaluation}
\subsection{Dataset}
One of the most common publicly available antipodal grasping dataset used to benchmark results is the Cornell Grasping Dataset [18]. The Cornell Grasping Dataset includes 885 RGB-D images of various real objects, with 5110 human-labelled positive and 2909 negative grasps.  The ground truth data has several grasp rectangles representing grasping possibilities for each object. In general, this is a fairly small dataset to use for training on the GGR-CNN model. To increase the size of the dataset and the accuracy of our model, we augment the original Cornell Grasping Dataset using random crops, rotations and zooms to create over 9000 depth images and the associated grasp map. Through augmentation, we can integrate over 63k grasp examples.

\subsection{Dataset Metric for Grasping Detection}
We detect a grasp using the rectangle representation. Based on this, a grasp is successfully detected when the difference between the grasp orientation of the predicted grasp rectangle and the ground truth rectangle is less than 30◦. Alongside this, the IoU score is between the ground truth grasp rectangle and the predicted grasp rectangle is more than 25\%. Since the model outputs an image-based grasp representation, the value corresponding to each pixel in the output image is mapped to its equivalent rectangle representation. 

\subsection{Cornell Grasping Results}
To evaluate the performance of our model on the Cornell Dataset, we have a cross-validation setup. We do an image-wise (IW) split, where the training set and the validation set do not share the same image and an object-wise (OW) data split, where the training set and the validation set do not share any images from the same object.

\begin{table}
  \caption{Results}
  \label{table4}
  \centering
  \begin{tabular}{lllll}           \\
    \cmidrule(r){1-5}
    Authors     & Model    & Accuracy (\%) - IW  & Accuracy (\%) - OW     & Speed (ms) \\
    \midrule
    Lenz [19] & SAE, struct. reg. & 73.9 & 75.6  & 1350   \\
    Redmon [20]  & AlexNet, MultiGrasp & 88.0 & 87.1   & 76   \\
    Kumra [16] & ResNet-50x2 & 89.2 & 88.9  & 103   \\
    Morrison [15] & GG-CNN & 73.0 & 69.0  & 19  \\
    Zhou [17] & FCGN, ResNet-101 & 97.7 & 96.6  & 117   \\
    Asif [21] & GraspNet & 90.2 & 90.6  & 24   \\
    Our model & GGR-CNN  & 92.4 & 91.1  &  22   \\
    \bottomrule
  \end{tabular}
\end{table}

\section{Application:  Open Arms in Teleoperated Nursing Robot Use}

In one experiment involving applications of the Open Arm project, we collaborated on the control of the arms with another team for a teleoperated nursing robot that we named Asha. The team developed an avatar system, through which a teleoperator can see, hear and interact with a remote environment.

\subsection{Teleoperation System Architecture}

\begin{figure}
  \centering
  \includegraphics[width=\linewidth]{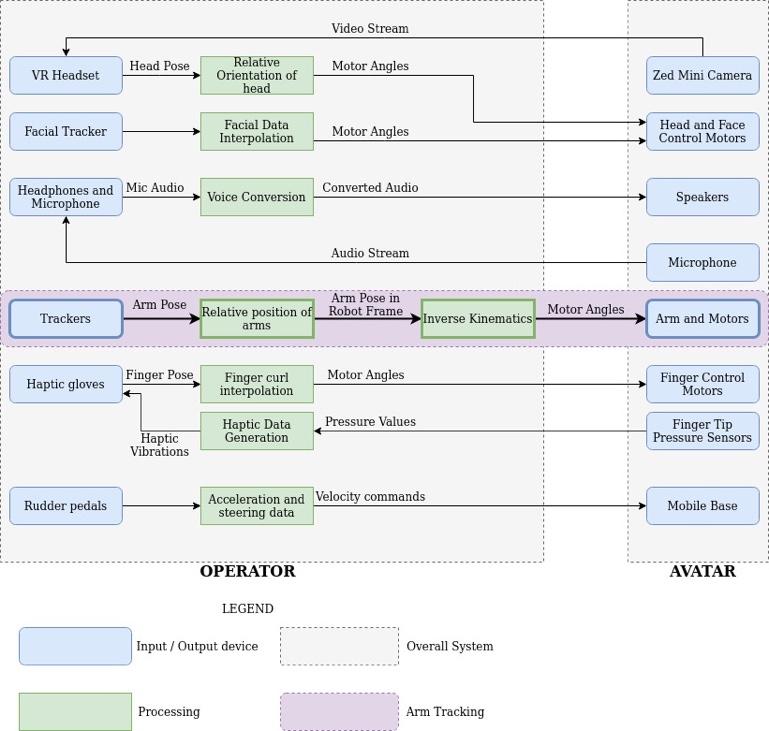}
  \caption{Teleoperation System Architecture}
\end{figure}

\textit{Audio and Video Feed to Operator} - The operator is equipped with a HTC Vive Pro Eye Virtual Reality (VR) headset which receives a video stream of the remote scene. The video stream is recorded by a Zed mini stereo camera that is mounted on Asha’s head. The VR headset enables the operator to perceive depth information and provides an immersive experience. A pair of headphones worn by the operator receive the audio stream recorded by a microphone array on the avatar side. 

\textit{Facial Expression Tracking} - A facial tracker module is mounted on the VR headset that determines the state of the lips/mouth of the operator. In addition to this, the VR headset measures the state of the eyes of the operator. The combination of this data is used to construct a facial expression that is then decomposed into a sequence of motor angles for each of the face motors on Asha. 

\textit{Avatar audio and Lip Control} - The operator’s speech is recorded using a microphone and then directly played back on the audio speakers of Asha. Parallely, viseme information is extracted from the processed speech and used to control the lip motors of the Asha to provide the appropriate lip sync.

\textit{Avatar Arms and Head Control} - In order to control the anthropomorphic Open Arms of Asha, the operator wears three HTC Vive VR trackers (two of them mounted on the back of each wrist and the third mounted on the operator’s chest). The trackers provides an estimate of the pose of the operator’s arm. A suitably designed Inverse Kinematics (IK) engine maps the operator’s arm pose to a corresponding arm pose for Asha. The VR headset also serves to estimate the head pose of the operator which can be used to control Asha’s head through the neck motors.

\textit{Haptic Feedback and Finger Tracking} - The Bebop gloves are used to provide a haptic response to the operator. It is possible to simulate haptic responses ranging from a light touch to a hard press, the strength of which is determined by the signals recorded by pressure sensors mounted on Asha’s fingertips.
The gloves are also equipped with flex sensors on each finger that records the degree to which a finger has curled. This allows for granular control of each individual finger and provides the ability to manipulate a variety of objects.

\textit{Navigation} - Asha is mounted on a differential drive Unmanned Ground Vehicle (UGV) to enable navigating the remote scene. Navigation commands are provided by the operator through a three pedal foot interface system (Logitech G27). 

\subsection{Inverse Kinematics Solvers for Accurate Teleoperation of Arms}
To be able to do useful tasks, the robot will have to track the arm movements of a remote operator accurately and smoothly. This can be done in two ways: 1) Replicate joint motions of the operator on the robot, 2) track the operator's hand movement and ensure the hand (i.e., the end-effector) of the robot follows the motion accurately.

Replicating the joint motions of the operator on the robot is possible in this particular case since Asha is anthromorphic in its build. Our approach is to track the movement of the operator's hand in 3D space by using a motion capture system and then using an inverse kinematics solver to derive the joint angles from the end-effector pose (position and orientation). The motion capture system that we are using is the HTC Vive. A tracker is placed on the back of the operator's palm and its motion is tracked within a pre-configured play area.

\subsection{Analytical vs Numerical Solvers}
Inverse kinematics solvers use either analytical solvers or numerical solvers to solve for the joint angles given the end-effector pose.

Analytical solvers such as IKFast [22] generate analytical expressions to map the joint angles to the end-effector poses. The expressions can be derived once and then subsequently used to solve the IK problem. Using these analytical expressions, the solutions can be obtained within a few microseconds which is orders of magnitude faster than numerical approaches. However, this solver requires that one joint in a 7-DoF manipulator be set as free so that there are only 6 unknown joint angles (which equals the 6 DoF of the end-effector). 

Numerical solvers, such as Kinematics and Dynamics Library (KDL) [24] and TRAC\_IK [25] use numerical optimization techniques to solve the inverse kinematics problem. These solvers can be orders of magnitude slower than the analytical solvers, but they provide other benefits. Numerical solvers can handle joint limits more gracefully by providing a best-fit solution that is close to the desired output, while analytical solvers do not provide an output if the constraints are not met. This is important because the range of motion of the joints of the operator need not always match the range of motion of the joints of the robot. 

\subsection{Evaluating Numerical Solvers: KDL vs. TRACK\_IK}
Two numerical solvers were evaluated for this work, namely, KDL and TRAC\_IK. The performance of the solvers was compared based on the solve rate. The solve rate is defined as the ratio of the number of successful solutions to the total number of test inputs given to a solver (expressed in percentage). A forward kinematics solver was used to find end-effector poses for random input joint angles and this end-effector pose was then provided as input to the inverse kinematics solvers. TRAC\_IK was able to find solutions for more than 99.8\% of all inputs whereas KDL was only able to solve 96\% of the test inputs. Based on these results TRAC\_IK was chosen as the IK solver in this work. 

\subsection{Improving Solve Rate}
A few methods can be used to improve the solve rate of the IK solver. 
Using a seed solution as an input to the solver helps to find the solution faster and it also ensures that the new solution is closer to the older one thus ensuring smooth arm motion of the robot. 

Using a two-stage approach improves the solve rate as the solvers will be able to find solutions easily when the tolerances are loose. In case the solver is unable to find solutions in stage 2 (with tighter tolerances), then the solution from stage 1 is sent to the robot. Although it is preferred to have solutions as close as possible to the commanded output, sending out approximate solutions without skipping commands ensures that the robot arms move smoothly.

\section{Conclusion}
Open Arms is a modular, open-source robotic arm and hand toolkit with mechanical capabilities comparable to the human hand and arm. The Open Arms platform has proven valuable and robust as an experimental platform designed to be mass-manufactured at a low cost. Our system includes touch sensors, force and position feedback in all joints, and fully integrated hardware with various cognitive AI controls, including our novel GGR-CNN model for accurate grasping control.

We hope that ongoing experiments on our social robot platform and robust simulation tools may produce new opportunities for experiments in embodied cognition and help progress towards more generalized intelligence in machines. 

\section{Contributions}
This work was initiated and primarily developed by the Hanson Robotics team to extend the capabilities of Sophia, a social humanoid robot developed by Hanson Robotics for AI and robotics research, applications development, and the arts. Gerardo Morales and David Hanson created the design and construction of the hands and arms as part of the Sophia robot platform. Alishba Imran focused on developing a generative grasping CNN method for improving the manipulation and grasping of the hands. In one experiment involving applications of the Sophia Open Arms project, Hanson Robotics collaborated on the control of the arms with the team at the Indian Institute of Science (IISc). In addition, Hanson Robotics team developed the 20th unit of Sophia Utility Platform with Open Arms and contributed the robot to a team including IISc, TCS, and TATA for a teleoperated nursing robot.
\section{References}
{
\small

[1] S. Funabashi, A. Schmitz, T. Sato, S. Somlor, and S. Sugano, “Versatile
in-hand manipulation of objects with different sizes and shapes using
neural networks,” in 2018 IEEE-RAS 18th International Conference
on Humanoid Robots (Humanoids), Nov 2018, pp. 1–9.

[2] OpenAI, “Learning dexterous in-hand manipulation,”
CoRR, vol. abs/1808.00177, 2018. [Online]. Available:
http://arxiv.org/abs/1808.00177.

[3] SimLab. Allegro hand overview, 2016. [Online; accessed 25-May-2020].

[4] Scott, C. (2018). Open Bionics Introduces the Hero Arm: First-Ever Medically Approved 3D Printed Bionic Arm. https://3dprint.com/208598/open-bionics-hero-arm/.

[5] S Park, H Lee, D Hanson, PY Oh, Sophia-Hubo's Arm Motion Generation for a Handshake and Gestures”, 15th International Conference on Ubiquitous Robots (UR), 2018 - ieeexplore.ieee.org.

[6] D. Hanson, A. Imran (Hanson Robotics), A. Vellanki, S. Kanagaraj,
A Neuro-Symbolic Humanlike Arm Controller for Sophia the Robot, Arxiv, 27 Oct 2020.

[7] Parmiggiani, Alberto \& Maggiali, Marco \& Natale, Lorenzo \& Nori, Francesco \& Schmitz, Alexander \& Tsagarakis, Nikos \& Santos-Victor, José \& Becchi, Francesco \& Sandini, Giulio \& Metta, Giorgio. (2012). The Design of the iCub Humanoid Robot. International Journal of Humanoid Robotics. 9.10.1142/S0219843612500272.

[8] Laird, John \& Kinkade, Keegan \& Mohan, Shiwali \& Xu, Joseph. (2012). Cognitive Robotics Using the Soar Cognitive Architecture. AAAI Workshop - Technical Report.

[9] Trafton, Greg \& Hiatt, Laura \& Harrison, Anthony \& Tanborello, Frank \& Khemlani, Sangeet \& Schultz, Alan. (2013). ACT-R/E: An embodied cognitive architecture for human-robot interaction. Journal of Human-Robot Interaction. 2. 30-55. 10.5898/JHRI.2.1.Trafton.

[10] Brooks, Rodney \& Breazeal, Cynthia \& Marjanovic, Matthew \& Scassellati, Brian \& Williamson, Matthew. (2002). The Cog Project: Building a Humanoid Robot. Lecture Notes in Artificial Intelligence. 1562. 10.1007/3-540-48834-0\_5.

[11] Lerrel Pinto and Abhinav Gupta. Supersizing selfsupervision: Learning to grasp from 50k tries and 700 robot hours. In Proc. of the IEEE International Conference on Robotics and Automation (ICRA), pages 3406–3413, 2016.

[12] Joseph Redmon and Anelia Angelova. Real-Time Grasp
Detection Using Convolutional Neural Networks. In Proc. of the IEEE International Conference on Robotics
and Automation (ICRA), pages 1316–1322, 2015.

[13] Ian Lenz, Honglak Lee, and Ashutosh Saxena. Deep learning for detecting robotic grasps. The International Journal of Robotics Research (IJRR), 34(4-5):705–724, 2015.

[14] Jeffrey Mahler, Jacky Liang, Sherdil Niyaz, Michael Laskey, Richard Doan, Xinyu Liu, Juan Aparicio Ojea, and Ken Goldberg. Dex-Net 2.0: Deep Learning to Plan Robust Grasps with Synthetic Point Clouds and Analytic Grasp Metrics. In Robotics: Science and Systems (RSS), 2017.

[15] Douglas Morrison, Peter Corke, and Jurgen Leitner. Learning robust, real-time, reactive robotic grasping. The International Journal of Robotics Research, page 0278364919859066, 2019.

[16] Sulabh Kumra and Christopher Kanan. Robotic grasp detection using deep convolutional neural networks. In 2017 IEEE/RSJ International Conference on Intelligent Robots and Systems (IROS), pages 769–776. IEEE, 2017.

[17] Xinwen Zhou, Xuguang Lan, Hanbo Zhang, Zhiqiang Tian, Yang Zhang, and Narming Zheng. Fully convolutional grasp detection network with oriented anchor box. In 2018 IEEE/RSJ International Conference on Intelligent Robots and Systems (IROS), pages 7223–7230. IEEE, 2018.

[18] Yun Jiang, Stephen Moseson, and Ashutosh Saxena. Efficient grasping from rgbd images: Learning using a new rectangle representation. In Robotics and Automation (ICRA), 2011 IEEE International Conference on, pages 3304–3311. IEEE, 2011

[19] Ian Lenz, Honglak Lee, and Ashutosh Saxena. Deep learning for detecting robotic grasps. The International Journal of Robotics Research, 34(4-5):705–724, 2015.

[20] Joseph Redmon and Anelia Angelova. Real-time grasp detection using convolutional neural networks. In 2015 IEEE International Conference on Robotics and Automation (ICRA), pages 1316–1322. IEEE, 2015.

[21] Umar Asif, Jianbin Tang, and Stefan Harrer. Graspnet: An efficient convolutional neural network for real time grasp detection for low-powered devices. In IJCAI, pages 4875–4882, 2018.

[22] Demonstration of Our Baseline Teleoperation System: https://drive.google.com/file/d/1EqFkXizNBdBass1-jc26gTMhr8gtqECd/view?usp=sharing

[23] Diankov, Rosen. "Automated construction of robotic manipulation programs." (2010).

[24] Smits, R." KDL: Kinematics and Dynamics Library", https://www.orocos.org/kdl

[25] Beeson, Patrick \& Ames, Barrett. (2015). TRAC-IK: An Open-Source Library for Improved Solving of Generic Inverse Kinematics. 10.1109/HUMANOIDS.2015.7363472.

}


\end{document}